\documentclass[runningheads]{llncs}
\usepackage[T1]{fontenc}

\usepackage[utf8]{inputenc} 
\usepackage[T1]{fontenc}    
\usepackage{url}            
\usepackage{booktabs}       
\usepackage{amsfonts}       
\usepackage{amssymb}
\usepackage{amsmath}
\usepackage{nicefrac}       
\usepackage{microtype}      
\usepackage[dvipsnames,table,xcdraw]{xcolor}        
\usepackage{array}
\usepackage{graphicx}
\usepackage{subcaption}
\usepackage{mwe,tikz}
\usepackage[percent]{overpic}
\usepackage{pgfplots}
\usepackage{multirow}
\usepackage{algorithmic}
\usepackage{algorithm}
\usepackage{verbatim} 

\usepackage{tikz-3dplot}

\usepackage[pagebackref,breaklinks,colorlinks]{hyperref}

\usetikzlibrary{plotmarks}
\usetikzlibrary{positioning,arrows.meta,quotes}
\usetikzlibrary{shapes,snakes}
\usetikzlibrary{bayesnet}
\usetikzlibrary{matrix}
\tikzset{>=latex}

\makeatletter
\newcommand{\HEADER}[1]{\ALC@it\underline{\textsc{#1}}\begin{ALC@g}}
\newcommand{\ENDHEADER}{\end{ALC@g}}
\newcommand{\STATEI}[1]{\STATE
  \begin{tabular}[t]{@{}p{\dimexpr \textwidth-\labelwidth-\ALC@tlm}@{}}%
    \hangindent \algorithmicindent
    \hangafter 1
    #1
  \end{tabular}
}
\makeatother

\usepackage[capitalize]{cleveref}
\crefname{section}{Sec.}{Secs.}
\Crefname{section}{Section}{Sections}
\Crefname{table}{Table}{Tables}
\crefname{table}{Tab.}{Tabs.}

\newcolumntype{L}[1]{>{\raggedright\let\newline\\\arraybackslash\hspace{0pt}}m{#1}}
\newcolumntype{C}[1]{>{\centering\let\newline\\\arraybackslash\hspace{0pt}}m{#1}}
\newcolumntype{R}[1]{>{\raggedleft\let\newline\\\arraybackslash\hspace{0pt}}m{#1}}

\DeclareMathOperator*{\argmax}{arg\,max}

\begin{document}

\title{GriTS: Grid table similarity metric for table structure recognition}
\titlerunning{GriTS: Grid table similarity}

\author{Brandon Smock\inst{1}\orcidID{0009-0002-7002-0800} \and
Rohith Pesala\inst{1}\orcidID{0009-0004-7373-853X} \and
Robin Abraham\inst{1}\orcidID{0000-0003-1915-8118}}

\authorrunning{B. Smock et al.}

\institute{Microsoft, Redmond WA, USA\\
\email{\{brsmock,ropesala,robin.abraham\}@microsoft.com}}

\maketitle

\begin{abstract}
In this paper, we propose a new class of metric for table structure recognition (TSR) evaluation, called grid table similarity (GriTS).
Unlike prior metrics, GriTS evaluates the correctness of a predicted table directly in its natural form as a matrix.
To create a similarity measure between matrices, we generalize the two-dimensional largest common substructure (2D-LCS) problem, which is NP-hard, to the 2D most similar substructures (2D-MSS) problem and propose a polynomial-time heuristic for solving it.
This algorithm produces both an upper and a lower bound on the true similarity between matrices.
We show using evaluation on a large real-world dataset that in practice there is almost no difference between these bounds.
We compare GriTS to other metrics and empirically validate that matrix similarity exhibits more desirable behavior than alternatives for TSR performance evaluation.
Finally, GriTS unifies all three subtasks of cell topology recognition, cell location recognition, and cell content recognition within the same framework, which simplifies the evaluation and enables more meaningful comparisons across different types of TSR approaches.
Code will be released at \url{https://github.com/microsoft/table-transformer}.
\end{abstract}

\section{Introduction}

Table extraction (TE) \cite{pinto2003table,yildiz2005pdf2table,correa2017unleashing} is the problem of inferring the presence, structure, and---to some extent---meaning of tables in documents or other unstructured presentations.
In its presented form, a table is typically expressed as a collection of cells organized over a two-dimensional grid \cite{wang1996tabular,gatterbauer2007towards,oro2009trex}. 
Table structure recognition (TSR) \cite{gobel2012methodology,gobel2013icdar} is the subtask of TE concerned with inferring this two-dimensional cellular structure from a table's unstructured presentation.

While straightforward to describe, formalizing the TSR task in a way that enables effective performance evaluation has proven challenging \cite{hassan2010towards}.
Perhaps the most straightforward way to measure performance is to compare the sets of predicted and ground truth cells for each table and measure the percentage of tables for which these sets match exactly---meaning, for each predicted cell there is a matching ground truth cell, and vice versa, that has the same rows, columns, and text content.
However, historically this metric has been eschewed in favor of measures of \emph{partial} correctness that score each table's correctness on a range of $[0, 1]$ rather than as binary correct or incorrect.
Measures of partial correctness are useful not only because they are more granular, but also because they are less impacted by errors and ambiguities in the ground truth.
This is important, as creating unambiguous ground truth for TSR is a challenging problem, which can introduce noise not only into the learning task but also performance evaluation \cite{smock2022pubtables,hu2001table}.

\usetikzlibrary{matrix, positioning, patterns}
\tikzstyle{colstyle} = []  
\tikzstyle{rowstyle} = []
\tikzstyle{Bstyle} = [fill=teal!30]

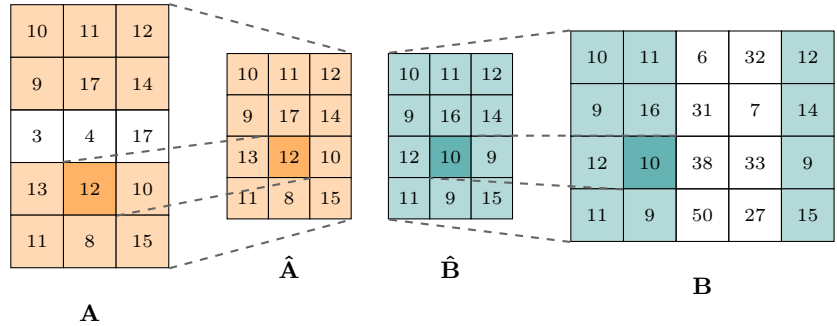
\begin{figure*}[t]
\centering
\begin{tikzpicture}[
    2d-arr/.style={matrix of nodes, row sep=-\pgflinewidth, column sep=-\pgflinewidth, nodes={draw, minimum width=0.7cm,
    minimum height=0.7cm}}
  ]
  \scriptsize
  \matrix (A) [2d-arr] {
  |[fill=orange!30]| 10 & |[fill=orange!30]| 11 & |[fill=orange!30]| 12 \\
  |[fill=orange!30]| 9 & |[fill=orange!30]| 17 & |[fill=orange!30]| 14 \\
  |[colstyle]| 3 & |[colstyle]| 4 & |[colstyle]| 17 \\
  |[fill=orange!30]| 13 & |[fill=orange!60]| 12 & |[fill=orange!30]| 10 \\
  |[fill=orange!30]| 11 & |[fill=orange!30]| 8 & |[fill=orange!30]| 15 \\
  };

  \node[below=1em of A] {\small $\mathbf A$};

  \matrix (AA) [2d-arr, right=2em of A, nodes={draw, fill=orange!30, minimum width=0.55cm,
    minimum height=0.55cm}] {
    10 & 11 & 12 \\
    9 & 17 & 14 \\
    13 & |[fill=orange!60]| 12 & 10 \\
    11 & 8 & 15 \\
  };
  \node[below=1em of AA] {\small $\mathbf{\hat{A}}$};

  \matrix (BB) [2d-arr, right=1em of AA, nodes={draw, fill=teal!30,
  minimum width=0.55cm, minimum height=0.55cm}] {
    10 & 11 & 12 \\
    9 & 16 & 14 \\
    12 & |[fill=teal!60]| 10 & 9 \\
    11 & 9 & 15 \\
  };
  \node[below=1em of BB] {\small $\mathbf{\hat{B}}$};

  \matrix (B) [2d-arr, right=2em of BB] {
  |[Bstyle]| 10 & |[Bstyle]| 11 & 6 & 32 & |[Bstyle]| 12\\
  |[Bstyle]| 9 & |[Bstyle]| 16 & 31 & 7 & |[Bstyle]| 14\\
  |[Bstyle]| 12 & |[fill=teal!60]| 10 & 38 & 33 & |[Bstyle]| 9\\
  |[Bstyle]| 11 & |[Bstyle]| 9 & 50 & 27 & |[Bstyle]| 15\\
  };
  \node[below=1em of B] {\small $\mathbf B$};

  \draw[dashed, black!60, thick] (A-1-3.north east) -- (AA-1-3.north east);
  \draw[dashed, black!60, thick] (A-4-2.north west) -- (AA-3-2.north west);
  \draw[dashed, black!60, thick] (A-5-3.south east) -- (AA-4-3.south east);
  \draw[dashed, black!60, thick] (A-4-2.south east) -- (AA-3-2.south east);

  \draw[dashed, black!60, thick] (B-3-2.north east) -- (BB-3-2.north east);
  \draw[dashed, black!60, thick] (B-1-1.north west) -- (BB-1-1.north west);
  \draw[dashed, black!60, thick] (B-3-2.south west) -- (BB-3-2.south west);
  \draw[dashed, black!60, thick] (B-4-1.south west) -- (BB-4-1.south west);
\end{tikzpicture}
\caption{Grid table similarity (GriTS) is based on computing the two-dimensional most similar substructures (2D-MSS) between two matrices (see \cref{sec:metrics} for details). In this example, the 2D-MSS of matrices $\mathbf A$ and $\mathbf B$ are substructures $\mathbf{\hat{A}}$ and $\mathbf{\hat{B}}$. $\mathbf{\hat{A}}$ and $\mathbf{\hat{B}}$ create a direct correspondence (or alignment) between entries in the original matrices, depicted here as darker shading for one pair of corresponding entries.}
\end{figure*}

Designing a metric for partial correctness of tables has also proven challenging.
The naive approach of comparing predicted cells with ground truth cells by their absolute positions suffers from the problem that a single mistake in cell segmentation can offset all subsequent cells by one position, which may result in a disproportionate penalty.
Several metrics have been proposed that instead consider the \emph{relative} positions of predicted cells \cite{gobel2012methodology,gao2019icdar,zhong2019image,li2020tablebank}.
However, these metrics capture relative position in different ways that do not fully account for a table's global two-dimensional (2D) structure.
Metrics also offer differing perspectives on what constitutes the task to be measured, what property of a predicted cell is evaluated, and whether predicted cells can be partially correct.

To address these issues, in this paper we develop a new class of metric for TSR called grid table similarity (GriTS).
GriTS attempts to unify the different perspectives on TSR evaluation and address these in a modular way, both to simplify the evaluation and to make comparisons between approaches more meaningful.
Among our contributions:
\begin{itemize}
  \item GriTS is the first metric to evaluate tables directly in their matrix form, maintaining the global 2D relationships between cells when comparing predictions to ground truth.
  \item To create a similarity between matrices, we extend the 2D largest common substructure (2D-LCS) problem, which is NP-hard, to 2D most similar substructures (2D-MSS) and propose a polynomial-time heuristic to solve it. This algorithm produces both an upper and lower bound on its approximation, which we show have little difference in practice.
  \item We outline the properties of an ideal TSR metric and validate empirically on a large real-world dataset that GriTS exhibits more ideal behavior than alternatives for TSR evaluation.
  \item GriTS is the first metric that addresses cell topology, cell content, and cell location recognition in a unified manner. This makes it easier to interpret and compare results across different modeling approaches and datasets.
\end{itemize}

\section{Related Work}

\begin{table*}[]
  \footnotesize
  \caption{Comparison of metrics proposed for table structure recognition.}
  \label{tab:tsr_metrics}
  \centering
  \begin{tabular}{L{0.17\textwidth}L{0.16\textwidth}L{0.26\textwidth}L{0.27\textwidth}L{0.1\textwidth}}
    \toprule
    \textbf{Name} & \textbf{Task/Cell Property} & \textbf{Data Structure} & \textbf{Cell Partial \mbox{Correctness}} & \textbf{Form} \\
    \midrule
    $\textrm{DAR}_\textrm{Con}$ \cite{gobel2012methodology} & Content & Set of adjacency relations & Exact match & F-score \\
    $\textrm{DAR}_\textrm{Loc}$ \cite{gao2019icdar} & Location & Set of adjacency relations & Avg. at multiple IoU thresholds & F-score \\
    BLEU-4 \cite{li2020tablebank} & Topology \& function & Sequence of HTML tokens & Exact match & BLEU-4 \\
    TEDS \cite{zhong2019image} & Content \& function & Tree of HTML tags & Normalized Levenshtein similarity & TEDS \\ 
    \midrule
    $\textbf{GriTS}_\textbf{Top}$ & Topology & Matrix of cells & IoU & F-score \\
    $\textbf{GriTS}_\textbf{Con}$ & Content & Matrix of cells & Normalized LCS & F-score \\
    $\textbf{GriTS}_\textbf{Loc}$ & Location & Matrix of cells & IoU & F-score \\
    \midrule
  \end{tabular}
\end{table*}

A number of metrics exist for evaluating table structure recognition methods.
These include the cell adjacency-based metric used in the ICDAR 2013 Table Competition, which we refer to as directed adjacency relations (DAR) \cite{gobel2012methodology,gao2019icdar}, 4-gram BLEU score (BLEU-4) \cite{li2020tablebank}, and tree edit distance similarity (TEDS) \cite{zhong2019image}.
In \cref{tab:tsr_metrics} we categorize these metrics across four dimensions: subtask (cell property), data structure, cell partial correctness, and overall score formulation.

\subsection{Subtask/property} Each metric typically poses the table structure recognition task more specifically as one of the following:
\begin{enumerate}
  \item \textit{Cell topology recognition} considers the layout of the cells, specifically the rows and columns each cell occupies over a two-dimensional grid.
  \item \textit{Cell content recognition} considers the layout of cells and the text content of each cell.
  \item \textit{Cell location recognition} considers the layout of cells and the absolute coordinates of each cell within a document.
\end{enumerate}
One way to characterize these subtasks is in terms of the \emph{property} of the cell that is considered most central to the recognition task.
For cell topology, this can be considered the colspan and rowspan of each cell.
Each perspective is useful.
Cell content recognition is most aligned with the end goal of TE, but for table image input it can depend on the quality of optical character recognition (OCR).
Cell location recognition does not depend on OCR, but not every TSR method reports cell locations.
Cell topology recognition is independent of OCR and is applicable to all TSR methods, but is not anchored to the actual content of the cells by either text or location.
Thus, accurate cell topology recognition is necessary but not sufficient for successful TSR.

Functional analysis is the subtask of table extraction concerned with determining whether each cell is a key (header) or value.
While it is usually considered separate from structure recognition, metrics sometimes evaluate aspects of TSR and functional analysis jointly, such as TEDS and BLEU-4.

\subsection{Data structure}
A table is presented in two dimensions, giving it a natural representation as a grid or matrix of cells.
The objective of TSR is usually considered to be inferring this grid structure.
However, for comparing predictions with ground truth, prior metrics have used alternative abstract representations for tables.
These possibly decompose a table into sub-units other than cells and represent them in an alternate structure with relationships between elements that differ from those of a matrix.
The metrics proposed by G{\"o}bel et al. \cite{gobel2012methodology} and Gao et al. \cite{gao2019icdar} deconstruct the grid structure of a table into a \emph{set} of directed adjacency relations, corresponding to pairs of neighboring cells.
We refer to the first metric, which evaluates cell text content, as $\text{DAR}_\text{Con}$, and the second, which evaluates cell location, as $\text{DAR}_\text{Loc}$.
Li et al. \cite{li2020tablebank} represent a table as a token \emph{sequence}, using a simplified HTML encoding.
Zhong et al. \cite{zhong2019image} also represent a table using HTML tokens but use the nesting of the tags to consider a table's cells to be \emph{tree}-structured, which more closely represents a table's two-dimensional structure than does a sequence.

Metrics based on these different representations each have their own sensitivities and invariances to changes to a table's structure.
Zhong et al. \cite{zhong2019image} investigate a few of these sensitivities and demonstrate that DAR mostly ignores the insertion of contiguous blank cells into a table, while TEDS does not.
However, largely the sensitivities of these metrics that result from their different representations have not been studied.
This makes it more challenging to interpret and compare them.

\subsection{Cell partial correctness}
Each metric produces a score between 0 and 1 for each table.
For some metrics this takes into account simply the fraction of matching sub-units between a prediction and ground truth.
Some metrics also define a measure of correctness for each sub-unit, which is between 0 and 1.
For instance, TEDS incorporates the normalized Levenshtein similarity to allow the text content of an HTML tag to partially match the ground truth.
Defining partial correctness at the cell level is useful because it is less sensitive to minor discrepancies between a prediction and ground truth that may have little or no impact on table extraction quality.

\subsection{Form}
The form of a metric is the way in which the match between prediction and ground truth is aggregated over matching and non-matching sub-units.
DAR uses both precision and recall, which can be taken together to produce the standard F-score.
BLEU-4 treats the output of a table structure recognition model as a sequence and uses the 4-gram BLEU score to measure the degree of match between a predicted and ground truth sequence.
TEDS computes a modified tree edit distance, which is the cost of transforming partially-matching and non-matching sub-units between the tree representations of a predicted and ground truth table.

\subsection{Spreadsheet diffing}

A related problem to the one explored in this paper is the identification of differences or changes between two versions of a spreadsheet \cite{chambers2010sheetdiff,harutyunyan2012planted}.
In this case the goal is to classify cells between the two versions as either modified or unmodified, and possibly to generate a sequence of edit transformations that would convert one version of a spreadsheet into another.

\section{Grid Table Similarity}\label{sec:metrics}

To motivate the metrics proposed in this paper, we first introduce the following attributes that we believe an ideal metric for table structure recognition should exhibit:
\begin{enumerate}
	\item \emph{Task isolation}: the table structure recognition task is measured in isolation from other table extraction tasks (detection and functional analysis).
	\item \emph{Cell isolation}: a true positive according to the metric corresponds to exactly one predicted cell and one ground truth cell.
	\item \emph{Two-dimensional order preservation}\label{item:ordering} \cite{amir2008generalized}: For any two true positive cells, $\textrm{tp}_1$ and $\textrm{tp}_2$, the relative order in which they appear is the same in both dimensions in the predicted and ground truth tables. More specifically:
	\begin{enumerate}
	    \item The maximum true row of $\textrm{tp}_1$ < minimum true row of $\textrm{tp}_2$ $\iff$ the maximum predicted row of $\textrm{tp}_1$ < minimum predicted row of $\textrm{tp}_2$.
	    \item The maximum true column of $\textrm{tp}_1$ < minimum true column of $\textrm{tp}_2$ $\iff$ the maximum predicted column of $\textrm{tp}_1$ < minimum predicted column of $\textrm{tp}_2$.
	    \item The maximum true row of $\textrm{tp}_1$ = minimum true row of $\textrm{tp}_2$ $\iff$ the maximum predicted row of $\textrm{tp}_1$ = minimum predicted row of $\textrm{tp}_2$.
	    \item The maximum true column of $\textrm{tp}_1$ = minimum true column of $\textrm{tp}_2$ $\iff$ the maximum predicted column of $\textrm{tp}_1$ = minimum predicted column of $\textrm{tp}_2$.
	\end{enumerate}
	\item \emph{Row and column equivalence}: the metric is invariant to transposing the rows and columns of both a prediction and ground truth (i.e. rows and columns are of equal importance).
	\item \emph{Cell position invariance}: the credit given to a correctly predicted cell is the same regardless of its absolute row-column position.
\end{enumerate}

The first two attributes are considered in \cref{tab:tsr_metrics}.
In \cref{sec:experiments}, we test the last two properties for different proposed metrics.
While not essential, we note again that in practice we believe it is also useful for a TSR metric to define partial correctness for cells and to have the same general form for both cell content recognition and cell location recognition.
In the remainder of this section we describe a new class of evaluation metric that meets all of the above criteria.

\subsection{2D-LCS} We first note that Property \ref{item:ordering} is difficult to enforce for cells that can span multiple rows and columns.
To account for this, we instead consider the matrix of \emph{grid cells} of a table.
Exactly one grid cell occupies the intersection of each row and each column of a table.
Note that as a spanning cell occupies multiple grid cells, its text content logically repeats at every grid cell location that the cell spans.

To enforce Property \ref{item:ordering} for grid cells, we consider the generalization of the longest common subsequence (LCS) problem to two dimensions, which is called the two-dimensional largest (or longest) common substructure (2D-LCS) problem \cite{amir2008generalized}.
Let $\mathbf{M}[R, C]$ be a matrix with $R = [r_1,\dots,r_m]$ representing its rows and $C = [c_1,\dots,c_n]$ representing its columns.
Let $R'\mid R$ be a subsequence of rows of $R$, and $C'\mid C$ be a subsequence of columns of $C$.
Then a substructure $\mathbf{M'}\mid\mathbf{M}$ is such that,
$$\mathbf{M'}\mid\mathbf{M} = \mathbf{M}[R', C'].$$
2D-LCS operates on two matrices, $\mathbf{A}$ and $\mathbf{B}$, and determines the largest two-dimensional substructures, $\mathbf{\hat{A}}\mid\mathbf{A} = \mathbf{\hat{B}}\mid\mathbf{B}$, the two have in common.
In other words, 
\begin{align}
\textrm{2D-LCS}(\mathbf{A},\mathbf{B}) &= \argmax_{\mathbf{A}'\mid\mathbf{A},\mathbf{B}'\mid\mathbf{B}} {\sum_{i,j} f(\mathbf{A}'_{i,j}, \mathbf{B}'_{i,j})}\\
&= \mathbf{\hat{A}},\mathbf{\hat{B}},
\end{align}
where,
\[
f(e_1, e_2) =  
\begin{cases}
    1,& \text{if } e_1 = e_2\\
    0,              & \text{otherwise}
\end{cases}.
\]

\subsection{2D-MSS} While a solution to the 2D-LCS problem satisfies Property \ref{item:ordering} for grid cells, it assumes an exact match between matrix elements.
To let cells partially match, an extension to 2D-LCS is to relax the exact match constraint and instead determine the two most \textit{similar} two-dimensional substructures, $\mathbf{\tilde{A}}$ and $\mathbf{\tilde{B}}$.
We define this by replacing equality between two entries $\mathbf{A}_{i,j}$ and $\mathbf{B}_{k,l}$ with a more general choice of similarity function between them. In other words,
\begin{align}
\textrm{2D-MSS}_f(\mathbf{A},\mathbf{B}) &= \argmax_{\mathbf{A}'\mid\mathbf{A},\mathbf{B}'\mid\mathbf{B}} {\sum_{i,j} f(\mathbf{A}'_{i,j}, \mathbf{B}'_{i,j})}\\
&= \mathbf{\tilde{A}},\mathbf{\tilde{B}},
\end{align}
where,
\[
0 \leq f(e_1, e_2) \leq 1 \qquad \forall e_1,e_2.
\]

Taking inspiration from the standard F-score, we define a general similarity measure between two matrices based on this as,
\begin{align}\label{eq:2dmss_sim}
\tilde{S}_f(\mathbf{A}, \mathbf{B}) = \frac{2\sum_{i,j} f(\mathbf{\tilde{A}}_{i,j}, \mathbf{\tilde{B}}_{i,j})} {{|\mathbf{A}|} + {|\mathbf{B}|}},
\end{align}
where $|\mathbf{M}_{m \times n}| = m \cdot n$.

\subsection{Grid table similarity (GriTS)} Finally, to define a similarity between tables, we use \cref{eq:2dmss_sim} with a particular choice of similarity function and a particular matrix of entries to compare.
This has the general form,
\begin{align}\label{eq:grits}
\text{GriTS}_f(\mathbf{A}, \mathbf{B}) = \frac{2\sum_{i,j} f(\mathbf{\tilde{A}}_{i,j}, \mathbf{\tilde{B}}_{i,j})} {{|\mathbf{A}|} + {|\mathbf{B}|}},
\end{align}
where $\mathbf{A}$ and $\mathbf{B}$ now represent tables---matrices of grid cells---and $f$ is a similarity function between the grid cells' properties.
Interpreting \cref{eq:grits} as an F-score, then letting $\mathbf{A}$ be the ground truth matrix and $\mathbf{B}$ be the predicted matrix, we can also define the following quantities, which we interpret as recall and precision: $\text{GriTS-Rec}_f(\mathbf{A}, \mathbf{B}) = \frac{\sum_{i,j} f(\mathbf{\tilde{A}}_{i,j}, \mathbf{\tilde{B}}_{i,j})} {|\mathbf{A}|}$ and $\text{GriTS-Prec}_f(\mathbf{A}, \mathbf{B}) = \frac{\sum_{i,j} f(\mathbf{\tilde{A}}_{i,j}, \mathbf{\tilde{B}}_{i,j})} {|\mathbf{B}|}$.

A specific choice of grid cell property and the similarity function between them defines a particular GriTS metric.
We define three of these: $\text{GriTS}_\text{Top}$ for cell topology recognition,  $\text{GriTS}_\text{Con}$ for cell text content recognition, and  $\text{GriTS}_\text{Loc}$ for cell location recognition.
Each evaluates table structure recognition from a different perspective.

\tikzset{db/.style={fill=white!90!black}}
\tikzset{dr/.style={fill=white!90!black}}
\tikzset{dg/.style={fill=white!90!black}}

\begin{figure*}[t]
\centering
\begin{subfigure}[b]{\textwidth}
  \centering
  \includegraphics[height=4cm]{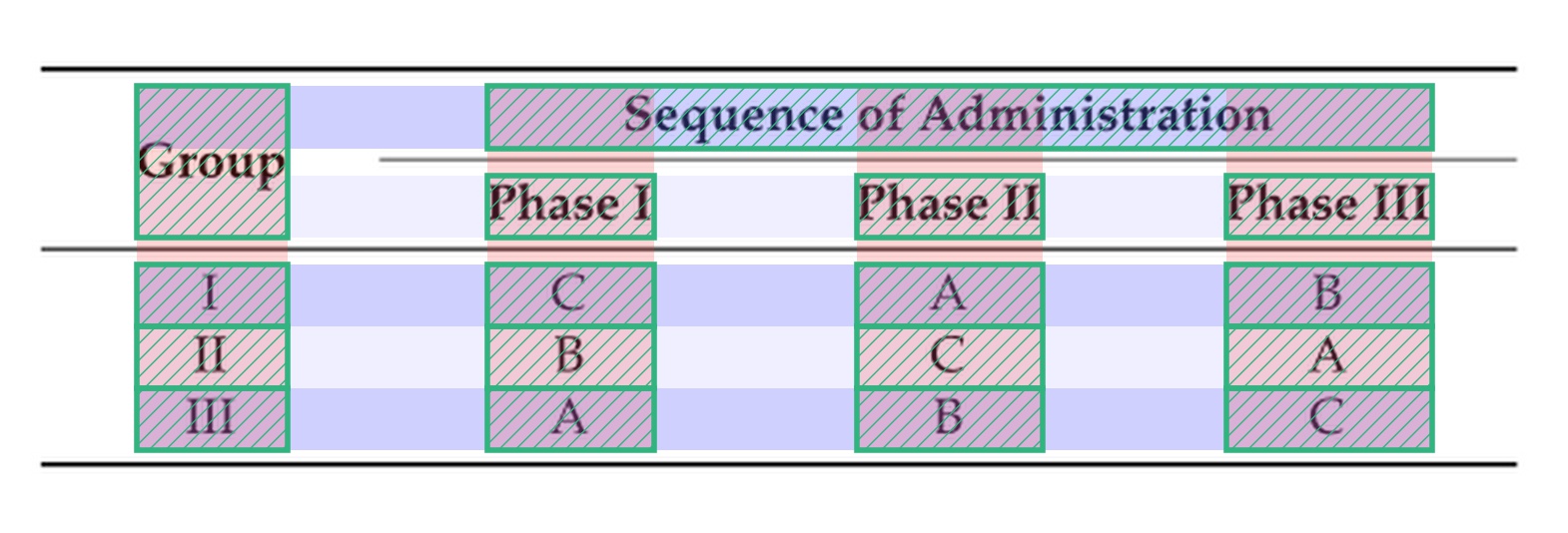}
	\caption{An example presentation table from the PubTables-1M dataset.\vspace{1em}}
	 \label{fig:presentation_table}
\end{subfigure}
\begin{subfigure}[b]{0.32\textwidth}
	\centering
	\begin{tikzpicture}[scale=0.4, every node/.style={scale=0.4}]
	\matrix[matrix of nodes,nodes={anchor=center, fill=white!95!white, draw=black!10!black,
                                                       text width=1.8cm, line width=0.01cm,
                                                       align=center, minimum width=2.2cm, minimum height=2cm, font=\LARGE}]{
		|[db]|\LARGE Group & |[db]|\scriptsize Sequence of Administration & |[db]|\scriptsize Sequence of Administration & |[db]|\scriptsize Sequence of Administration \\
		|[db]|\LARGE Group & \LARGE Phase I & \LARGE Phase II & \LARGE Phase III \\
		I & C & A & B \\
		II & B & C & A \\
		III & A & B & C \\};
	\end{tikzpicture}
	\caption{$\text{GriTS}_\text{Con}$}
	 \label{fig:grits_cont_matrix}
\end{subfigure}
\begin{subfigure}[b]{0.32\textwidth}
	\centering
	\begin{tikzpicture}[scale=0.4, every node/.style={scale=0.4}]
	\matrix[matrix of nodes,nodes={anchor=center, fill=white!94!white, draw=black!10!black,
                                                       text width=1.8cm, line width=0.01cm,
                                                       align=center, minimum width=2.2cm, minimum height=2cm, font=\Large}]{
		|[dr]|{[0, 0, 1, 2]} & |[dr]|{[0, 0, 3, 1]} & |[dr]|{[-1, 0, 2, 1]} & |[dr]|{[-2, 0, 1, 1]} \\
		|[dr]|{[0, -1, 1, 1]} & {[0, 0, 1, 1]} & {[0, 0, 1, 1]} & {[0, 0, 1, 1]} \\
		{[0, 0, 1, 1]} & {[0, 0, 1, 1]} & {[0, 0, 1, 1]} & {[0, 0, 1, 1]} \\
		{[0, 0, 1, 1]} & {[0, 0, 1, 1]} & {[0, 0, 1, 1]} & {[0, 0, 1, 1]} \\
		{[0, 0, 1, 1]} & {[0, 0, 1, 1]} & {[0, 0, 1, 1]} & {[0, 0, 1, 1]} \\};
	\end{tikzpicture}
	\caption{$\text{GriTS}_\text{Top}$}
	 \label{fig:grits_top_matrix}
\end{subfigure}
\begin{subfigure}[b]{0.32\textwidth}
	\centering
	\begin{tikzpicture}[scale=0.4, every node/.style={scale=0.4}]
	\matrix[matrix of nodes,nodes={anchor=center, fill=white!94!white, draw=black!10!black,
                                                       text width=1.8cm, line width=0.01cm,
                                                       align=center, minimum width=2.2cm, minimum height=2cm}]{
		|[dg]|{[136.42, 477.25, 160.62, 501.45]} & |[dg]|{[185, 477.25, 470.89, 487.22]} & |[dg]|{[185, 477.25, 470.89, 487.22]} & |[dg]|{[185, 477.25, 470.89, 487.22]} \\
		|[dg]|{[136.42, 477.25, 160.62, 501.45]} & {[185,  491.48, 271.9, 501.45]} & {[284.5, 491.48, 371.39, 501.45]} & {[384, 491.48, 470.89, 501.45]} \\
		{[136.42, 505.82, 160.62, 515.72]} & {[185, 505.82, 271.9, 515.72]} & {[284.5, 505.82, 371.39, 515.72]} & {[384, 505.82, 470.89, 515.72]} \\
		{[136.42, 515.73, 160.62, 525.63]} & {[185, 515.73, 271.9, 525.63]} & {[284.5, 515.73, 371.39, 525.63]} & {[384, 515.73, 470.89, 525.63]} \\
		{[136.42, 525.64, 160.62, 535.53]} & {[185, 525.64, 271.9, 535.53]} & {[284.5, 525.64, 371.39, 535.53]} & {[384, 525.64, 470.89, 535.53]} \\};
	\end{tikzpicture}
	\caption{$\text{GriTS}_\text{Loc}$}
	 \label{fig:grits_bbox_matrix}
\end{subfigure}
 \caption{An example presentation table from the PubTables-1M dataset, along with corresponding ground truth grid cell matrices for different GriTS metrics. Each matrix entry corresponds to one grid cell. Entries that correspond to spanning cells are shaded darker for illustrative purposes.}
 \label{fig:grits_matrices}
\end{figure*}

The matrices used for each metric are visualized in \cref{fig:grits_matrices}. 
For cell location, $\mathbf{A}_{i,j}$ contains the bounding box of the cell at row $i$, column $j$, and we use IoU to compute similarity between bounding boxes.
For cell text content, $\mathbf{A}_{i,j}$ contains the text content of the cell at row $i$, column $j$, and we use normalized longest common subsequence (LCS) to compute similarity between text sequences.

For cell topology, we use the same similarity function as cell location but on bounding boxes with size and relative position given in the grid cell coordinate system.
For the cell at row $i$, column $j$, let $\alpha_{i,j}$ be its rowspan, let $\beta_{i,j}$ be its colspan, let $\rho_{i,j}$ be the minimum row it occupies, and let $\theta_{i,j}$ be the minimum column it occupies.
Then for cell topology recognition, $\mathbf{A}_{i,j}$ contains the bounding box $[\theta_{i,j}-j, \rho_{i,j}-i, \theta_{i,j}-j+\beta_{i,j}, \rho_{i,j}-i+\alpha_{i,j}]$.
Note that for any cell with rowspan of 1 and colspan of 1, this box is $[0, 0, 1, 1]$.

\subsection{Factored 2D-MSS algorithm} Computing the 2D-LCS of two matrices is NP-hard \cite{amir2008generalized}.
This suggests that all metrics for TSR may necessarily be an approximation to what could be considered the ideal metric.
We propose a heuristic approach to determine the 2D-MSS by factoring the problem.
Instead of determining the optimal subsequences of rows and columns jointly for each matrix, we determine the optimal subsequences of rows and the optimal subsequences of columns independently.
This uses dynamic programming (DP) in a nested manner, which is run twice: once to determine the optimal subsequences of rows and a second time to determine the optimal subsequences of columns.
For the case of rows, an inner DP operates on sequences of grid cells in a row, computing the best possible sequence alignment of cells between any two rows.
The inner DP is executed over all pairs of predicted and ground truth rows to score how well each predicted row can be aligned with each ground truth row.
An outer DP operates on the sequences of rows from each matrix, using the pairwise scores computed by the inner DP to determine the best alignment of subsequences of rows between the predicted and ground truth matrices.
For the case of columns, the procedure is identical, merely substituting columns for rows.
The nested DP procedure is $O(|\mathbf{A}|\cdot|\mathbf{B}|)$.
Our implementation uses extensive memoization \cite{jaffar2008efficient} to maximize the efficiency of the procedure.

This factored procedure is similar to the RowColAlign algorithm \cite{harutyunyan2012planted} proposed for spreadsheet diffing.
Both procedures decouple the optimization of rows and columns and use DP in a nested manner.
However, RowColAlign attempts to optimize the number of pair-wise exact matches between substructures, whereas Factored 2D-MSS attempts to optimize the pair-wise similarity between substructures.
These differing objectives result in the two procedures having differing outcomes given the same two input matrices.

\subsection{Approximation bounds} The outcome of the procedure is a valid 2D substructure of each matrix---these just may not be the \emph{most similar} substructures possible, given that the rows and columns are optimized separately.
However, given that these are valid substructures, it follows that the similarity between matrices $\mathbf{A}$ and $\mathbf{B}$ computed by this procedure is a \emph{lower bound} on their true similarity.
It similarly follows that because constraints are relaxed during the optimization procedure, the lowest similarity determined when computing the optimal subsequences of rows and the optimal subsequences of columns serves as an \emph{upper bound} on the true similarity between matrices.
We define GriTS as the value of the lower bound, as it always corresponds to a valid substructure.
However, the upper bound score can also be reported to indicate if there is any uncertainty in the true value.
As we show in \cref{sec:experiments}, little difference is observed between these bounds in practice .

\section{Experiments}\label{sec:experiments}

\begin{table*}[]
  \footnotesize
  \caption{In this experiment we evaluate models of varying strengths on the PubTables-1M test set and compare the upper and lower bounds for $\textrm{GriTS}_\textrm{Con}$ that are produced. This shows how closely GriTS approximates the true similarity between two tables using the Factored 2D-MSS algorithm.}
  \label{tab:grits_bounds}
  \centering
  \begin{tabular}{L{0.1\textwidth}R{0.14\textwidth}R{0.21\textwidth}R{0.17\textwidth}R{0.3\textwidth}}
    \toprule
    \textbf{Epoch} & $\textbf{GriTS}_\textbf{Con}$ & \textbf{Upper Bound} & \textbf{Difference} & \textbf{Equal Instances (\%)} \\
    \midrule
    1 & 0.8795 & 0.8801 & 0.0005 & 81.2\% \\
    2 & 0.9347 & 0.9348 & 0.0002 & 87.6\% \\
    3 & 0.9531 & 0.9532 & 0.0001 & 91.1\% \\
    4 & 0.9640 & 0.9641 & < 0.0001 & 93.0\% \\
    5 & 0.9683 & 0.9683 & < 0.0001 & 93.2\% \\
    10 & 0.9794 & 0.9795 & < 0.0001 & 96.1\% \\
    15 & 0.9829 & 0.9829 & < 0.0001 & 96.9\% \\
    20 & 0.9850 & 0.9850 & < 0.0001 & 97.4\% \\
    \midrule
  \end{tabular}
\end{table*}

In this section, we report several experiments that assess and compare GriTS and other metrics for TSR.
Given that due to computational intractability no algorithm perfectly implements the ideal metric for TSR as outlined in \cref{sec:metrics}, the main goal of these experiments is to assess how well each proposed metric matches the behavior that we would expect in the theoretically optimal metric.

GriTS computes a similarity between two tables using the Factored 2D-MSS algorithm, which produces both an upper and a lower bound on the true similarity.
The difference between the two values represents the uncertainty, or maximum possible error, there is with respect to the true similarity.
In the first experiment, we measure how well GriTS approximates the true similarity between predicted and ground truth tables in practice.
To do this, we train the Table Transformer (TATR) model \cite{smock2022pubtables} on the PubTables-1M dataset, which contains nearly one million tables, for 20 epochs and save the model produced after each epoch.
This effectively creates 20 different TSR models of varying strengths with which to evaluate predictions.
We evaluate each model on the entire PubTables-1M test set and measure the difference between $\textrm{GriTS}_\textrm{Con}$ and $\textrm{GriTS}_\textrm{Con}$ upper bound, as well as the percentage of individual instances for which these bounds are equal.

We present the results of this experiment in \cref{tab:grits_bounds}.
As can be seen, there is very little difference between the upper and lower bounds across all models.
The uncertainty in the true table similarity peaks at the worst-performing model, which is trained for only one epoch.
For this model, $\textrm{GriTS}_\textrm{Con}$ is 0.8795 and the measured uncertainty in the true similarity is 0.0005.
Above a certain level of model performance, the average difference between the bounds and the percentage of instances for which the bounds differ both decrease quickly as performance improves, with the difference approaching 0 as GriTS approaches a score of 1. 
By epoch 4, the difference between the bounds is already less than 0.0001, and for at least 93\% of instances tested the GriTS score is in fact the true similarity.
These results strongly suggest that in practice there is almost no difference between GriTS and the true similarity between tables.

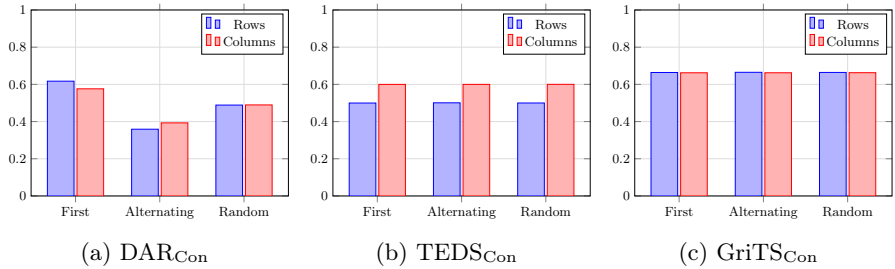
\begin{figure*}[t]
  \begin{subfigure}[b]{0.32\linewidth}
	\centering
	\resizebox{\linewidth}{!}{
	\begin{tikzpicture}
	\definecolor{light gray}{RGB}{220, 220, 220}
		\begin{axis} [grid, grid style={light gray, line cap=round}, ybar,enlarge x limits={abs=30pt},x=2cm,bar width=18pt,height=6cm,width=8cm,symbolic x coords={First, Alternating, Random},xtick = data, ymin=0, ymax=1]
		\addplot coordinates {
		    (First, 0.6172)
                    (Alternating, 0.3589)
                    (Random, 0.4887)
		};
		\addplot coordinates {
		    (First, 0.5761)
                    (Alternating,  0.3930)
                    (Random, 0.4895)
		};
		\legend {Rows, Columns};
		\end{axis}
	\end{tikzpicture}
	}
	\caption{$\text{DAR}_\text{Con}$}
    \label{subfig:metrics2.1}
  \end{subfigure}
  \begin{subfigure}[b]{0.32\linewidth}
	\centering
	\resizebox{\linewidth}{!}{
	\begin{tikzpicture}
	\definecolor{light gray}{RGB}{220, 220, 220}
		\begin{axis} [grid, grid style={light gray, line cap=round}, ybar,enlarge x limits={abs=30pt},x=2cm,bar width=18pt,height=6cm,width=8cm,symbolic x coords={First, Alternating, Random},xtick = data, ymin=0, ymax=1]
		\addplot coordinates {
		    (First, 0.4997)
                    (Alternating, 0.5008)
                    (Random, 0.4999)
		};
		\addplot coordinates {
		    (First, 0.5998)
                    (Alternating,  0.5999)
                    (Random, 0.6002)
		};
		\legend {Rows, Columns};
		\end{axis}
	\end{tikzpicture}
	}
	\caption{$\text{TEDS}_\text{Con}$}
    \label{subfig:metrics2.2}
  \end{subfigure}
  \begin{subfigure}[b]{0.32\linewidth}
	\centering
	\resizebox{\linewidth}{!}{
	\begin{tikzpicture}
	\definecolor{light gray}{RGB}{220, 220, 220}
		\begin{axis} [grid, grid style={light gray, line cap=round}, ybar,enlarge x limits={abs=30pt},x=2cm,bar width=18pt,height=6cm,width=8cm,symbolic x coords={First, Alternating, Random},xtick = data, ymin=0, ymax=1]
		\addplot coordinates {
		    (First, 0.6639)
                    (Alternating, 0.6649)
                    (Random, 0.6642)
		};
		\addplot coordinates {
		    (First, 0.6623)
                    (Alternating,  0.6624)
                    (Random, 0.6628)
		};
		\legend {Rows, Columns};
		\end{axis}
	\end{tikzpicture}
	}
	\caption{$\text{GriTS}_\text{Con}$}
    \label{subfig:metrics2.3}
  \end{subfigure}
	\caption{In this experiment, we compare the response of each metric to a predicted table with half of its rows missing (blue) or half of its columns missing (red) under different schemes for creating missing rows and columns. Results are averaged over 44,381 tables from the test set of PubTables-1M.}
	\label{fig:metrics2}
\end{figure*}

\begin{figure*}[t]
  \begin{subfigure}[b]{0.49\linewidth}
	\centering
	\resizebox{\linewidth}{!}{
	\begin{tikzpicture}[scale=1]
		\definecolor{light gray}{RGB}{220, 220, 220}
		\begin{axis}[grid, grid style={light gray, line cap=round}, xmin=0, xmax=1, ymin=0, ymax=1, legend pos=south east,
				   xlabel={$\textrm{p}(x)$}, ylabel={Score},
				   title={Rows}]
		\addplot[mark=none, light gray, line cap=round, forget plot] coordinates {(0,0) (1,1)};
		\addplot [blue, thick, line cap=round, mark=o, dashed, mark options={solid}, mark size=4pt]
			table {grits_top_row_p.dat};
		\addlegendentry[]{$\text{GriTS}_\text{Top}$}
		\addplot [red, thick, line cap=round, mark=o, dashdotted, mark size=3pt]
			table {grits_cont_row_p.dat};
		\addlegendentry[]{$\text{GriTS}_\text{Con}$}
		\addplot [orange, thick, line cap=round, mark=o, dotted, mark size=2pt]
			table {grits_loc_row_p.dat};
		\addlegendentry[]{$\text{GriTS}_\text{Loc}$}
		\addplot [green, thick, line cap=round, mark=square, dashed, mark options={solid},
		mark size=4pt]
			table {adjacency_fscore_cont_row_p.dat};
		\addlegendentry[]{$\text{DAR}_\text{Con}$}
		\addplot [magenta, thick, line cap=round, mark=triangle, dashed, mark options={solid},
		mark size=4pt]
			table {teds_row_p.dat};
		\addlegendentry[]{$\text{TEDS}_\text{Con}$}
		\end{axis}
	\end{tikzpicture}
	}
    \caption{}
    \label{subfig:metrics.2a}
  \end{subfigure}
  \begin{subfigure}[b]{0.49\linewidth}
	\centering
	\resizebox{\linewidth}{!}{
	\begin{tikzpicture}[scale=1]
		\definecolor{light gray}{RGB}{220, 220, 220}
		\begin{axis}[grid, grid style={light gray, line cap=round}, xmin=0, xmax=1, ymin=0, ymax=1, legend pos=south east,
				   xlabel={$\textrm{p}(x)$}, ylabel={Score},
				   title={Columns}]
		\addplot[mark=none, light gray, line cap=round, forget plot] coordinates {(0,0) (1,1)};
		\addplot [blue, thick, line cap=round, mark=o, dashed, mark options={solid}, mark size=4pt]
			table {grits_top_column_p.dat};
		\addlegendentry[]{$\text{GriTS}_\text{Top}$}
		\addplot [red, thick, line cap=round, mark=o, dashdotted, mark size=3pt]
			table {grits_cont_column_p.dat};
		\addlegendentry[]{$\text{GriTS}_\text{Con}$}
		\addplot [orange, thick, line cap=round, mark=o, dotted, mark size=2pt]
			table {grits_loc_column_p.dat};
		\addlegendentry[]{$\text{GriTS}_\text{Loc}$}
		\addplot [green, thick, line cap=round, mark=square, dashed, mark options={solid},
		mark size=4pt]
			table {adjacency_fscore_cont_column_p.dat};
		\addlegendentry[]{$\text{DAR}_\text{Con}$}
		\addplot [magenta, thick, line cap=round, mark=triangle, dashed, mark options={solid},
		mark size=4pt]
			table {teds_column_p.dat};
		\addlegendentry[]{$\text{TEDS}_\text{Con}$}
		\end{axis}
	\end{tikzpicture}
	}
    \caption{}
    \label{subfig:metrics.2b}
  \end{subfigure}
	\caption{In this experiment, we compare the response of each metric to a predicted table with random rows missing (\cref{subfig:metrics.2a}) or random columns missing (\cref{subfig:metrics.2b}) as we vary the probability that a row or column is missing. Results are averaged over 44,381 tables from the test set of PubTables-1M.}
	\label{fig:metrics}
\end{figure*}
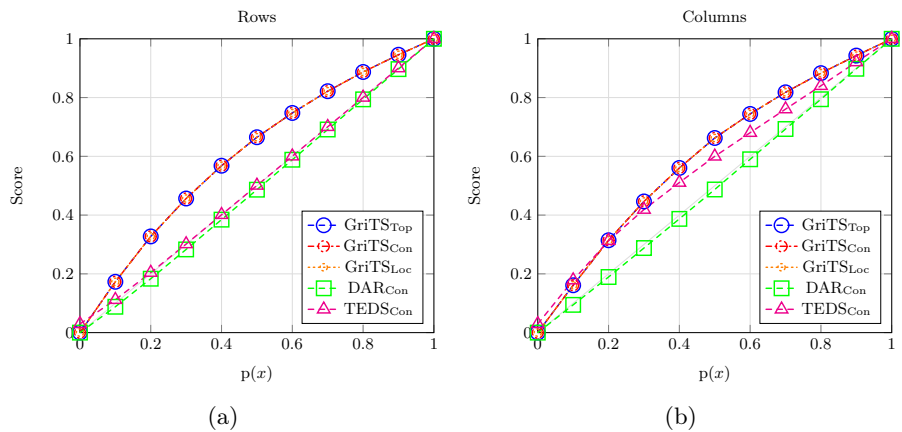

\begin{figure*}[t]
\centering
  \begin{subfigure}[b]{0.49\linewidth}
	\centering
	\resizebox{\linewidth}{!}{
	\begin{tikzpicture}[scale=1]
		\definecolor{light gray}{RGB}{220, 220, 220}
		\begin{axis}[grid, grid style={light gray, line cap=round}, xmin=0, xmax=1, ymin=0, ymax=1, legend pos=south east,
				   xlabel={$\textrm{p}(x)$}, ylabel={Recall},
				   title={Rows}]
		\addplot[mark=none, light gray, line cap=round, forget plot] coordinates {(0,0) (1,1)};
		\addplot [blue, thick, line cap=round, mark=o, dashed, mark options={solid}, mark size=4pt]
			table {grits_recall_top_row_p.dat};
		\addlegendentry[]{$\text{GriTS}_\text{Top}\text{ Rec.}$}
		\addplot [red, thick, line cap=round, mark=o, dashdotted, mark size=3pt]
			table {grits_recall_cont_row_p.dat};
		\addlegendentry[]{$\text{GriTS}_\text{Con}\text{ Rec.}$}
		\addplot [orange, thick, line cap=round, mark=o, dotted, mark size=2pt]
			table {grits_recall_loc_row_p.dat};
		\addlegendentry[]{$\text{GriTS}_\text{Loc}\text{ Rec.}$}
		\addplot [green, thick, line cap=round, mark=square, dashed, mark options={solid},
		mark size=4pt]
			table {adjacency_recall_cont_row_p.dat};
		\addlegendentry[]{$\text{DAR}_\text{Con}\text{ Rec.}$}
		\end{axis}
	\end{tikzpicture}
	}
    \label{subfig:pr.metrics.a}
  \end{subfigure}
  \begin{subfigure}[b]{0.49\linewidth}
	\centering
	\resizebox{\linewidth}{!}{
	\begin{tikzpicture}[scale=1]
		\definecolor{light gray}{RGB}{220, 220, 220}
		\begin{axis}[grid, grid style={light gray, line cap=round}, xmin=0, xmax=1, ymin=0, ymax=1, legend pos=south east,
				   xlabel={$\textrm{p}(x)$}, ylabel={Recall},
				   title={Columns}]
		\addplot[mark=none, light gray, line cap=round, forget plot] coordinates {(0,0) (1,1)};
		\addplot [blue, thick, line cap=round, mark=o, dashed, mark options={solid}, mark size=4pt]
			table {grits_recall_top_column_p.dat};
		\addlegendentry[]{$\text{GriTS}_\text{Top}\text{ Rec.}$}
		\addplot [red, thick, line cap=round, mark=o, dashdotted, mark size=3pt]
			table {grits_recall_cont_column_p.dat};
		\addlegendentry[]{$\text{GriTS}_\text{Con}\text{ Rec.}$}
		\addplot [orange, thick, line cap=round, mark=o, dotted, mark size=2pt]
			table {grits_recall_loc_column_p.dat};
		\addlegendentry[]{$\text{GriTS}_\text{Loc}\text{ Rec.}$}
		\addplot [green, thick, line cap=round, mark=square, dashed, mark options={solid},
		mark size=4pt]
			table {adjacency_recall_cont_column_p.dat};
		\addlegendentry[]{$\text{DAR}_\text{Con}\text{ Rec.}$}
		\end{axis}
	\end{tikzpicture}
	}
    \label{subfig:pr.metrics.b}
  \end{subfigure}
  \begin{subfigure}[b]{0.49\linewidth}
	\centering
	\resizebox{\linewidth}{!}{
	\begin{tikzpicture}[scale=1]
		\definecolor{light gray}{RGB}{220, 220, 220}
		\begin{axis}[grid, grid style={light gray, line cap=round}, xmin=0, xmax=1, ymin=0, ymax=1, legend pos=south east,
				   xlabel={$\textrm{p}(x)$}, ylabel={Precision},
				   title={Rows}]
		\addplot[mark=none, light gray, line cap=round, forget plot] coordinates {(0,0) (1,1)};
		\addplot [blue, thick, line cap=round, mark=o, dashed, mark options={solid}, mark size=4pt]
			table {grits_precision_top_row_p.dat};
		\addlegendentry[]{$\text{GriTS}_\text{Top}\text{ Prec.}$}
		\addplot [red, thick, line cap=round, mark=o, dashdotted, mark size=3pt]
			table {grits_precision_cont_row_p.dat};
		\addlegendentry[]{$\text{GriTS}_\text{Con}\text{ Prec.}$}
		\addplot [orange, thick, line cap=round, mark=o, dotted, mark size=2pt]
			table {grits_precision_loc_row_p.dat};
		\addlegendentry[]{$\text{GriTS}_\text{Loc}\text{ Prec.}$}
		\addplot [green, thick, line cap=round, mark=square, dashed, mark options={solid},
		mark size=4pt]
			table {adjacency_precision_cont_row_p.dat};
		\addlegendentry[]{$\text{DAR}_\text{Con}\text{ Prec.}$}
		\end{axis}
	\end{tikzpicture}
	}
    \label{subfig:pr.metrics.c}
  \end{subfigure}
  \begin{subfigure}[b]{0.49\linewidth}
	\centering
	\resizebox{\linewidth}{!}{
	\begin{tikzpicture}[scale=1]
		\definecolor{light gray}{RGB}{220, 220, 220}
		\begin{axis}[grid, grid style={light gray, line cap=round}, xmin=0, xmax=1, ymin=0, ymax=1, legend pos=south east,
				   xlabel={$\textrm{p}(x)$}, ylabel={Precision},
				   title={Columns}]
		\addplot[mark=none, light gray, line cap=round, forget plot] coordinates {(0,0) (1,1)};
		\addplot [blue, thick, line cap=round, mark=o, dashed, mark options={solid}, mark size=4pt]
			table {grits_precision_top_column_p.dat};
		\addlegendentry[]{$\text{GriTS}_\text{Top}\text{ Prec.}$}
		\addplot [red, thick, line cap=round, mark=o, dashdotted, mark size=3pt]
			table {grits_precision_cont_column_p.dat};
		\addlegendentry[]{$\text{GriTS}_\text{Con}\text{ Prec.}$}
		\addplot [orange, thick, line cap=round, mark=o, dotted, mark size=2pt]
			table {grits_precision_loc_column_p.dat};
		\addlegendentry[]{$\text{GriTS}_\text{Loc}\text{ Prec.}$}
		\addplot [green, thick, line cap=round, mark=square, dashed, mark options={solid},
		mark size=4pt]
			table {adjacency_precision_cont_column_p.dat};
		\addlegendentry[]{$\text{DAR}_\text{Con}\text{ Prec.}$}
		\end{axis}
	\end{tikzpicture}
	}
    \label{subfig:pr.metrics.d}
  \end{subfigure}
	\caption{In this experiment, we compare the response of each metric split into precision and recall to a predicted table with random rows missing or random columns missing as we vary the probability that a row or column is missing. Results are averaged over 44,381 tables from the test set of PubTables-1M.}
	\label{fig:metrics3}
\end{figure*}
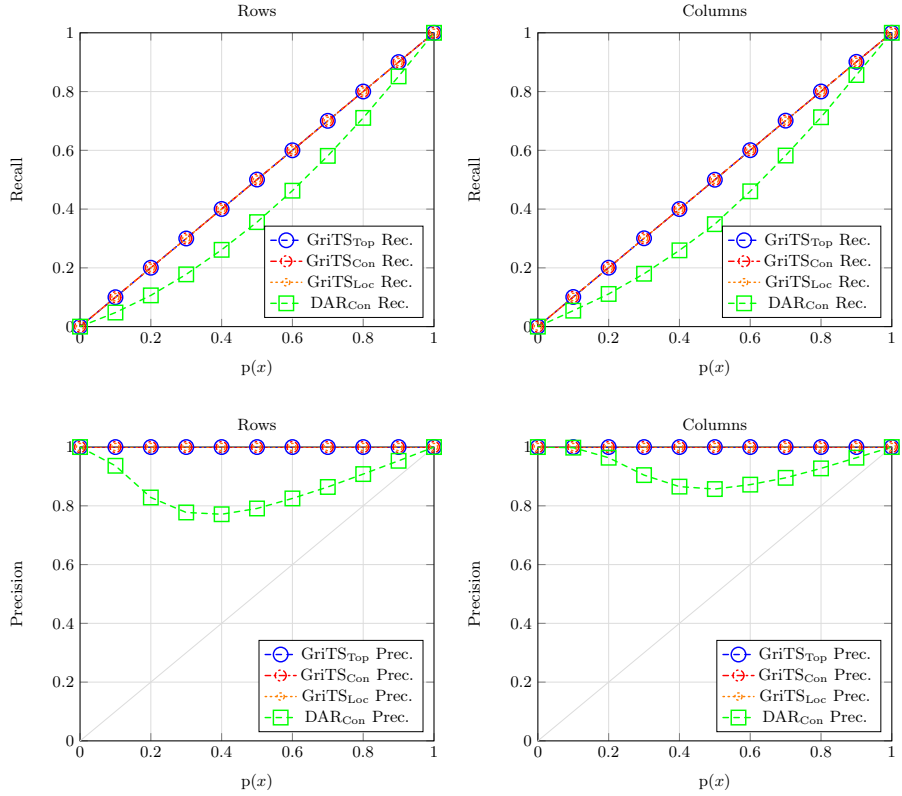

In the next set of experiments, we compare GriTS and other metrics for TSR with respect to the properties outlined in \cref{sec:metrics}.
The goal of these experiments is to assess how well each metric matches the behavior that we would expect in the theoretically optimal metric.
We evaluate each metric on the original ground truth (GT) and versions of the ground truth that are modified or corrupted in controlled ways, which shows the sensitivity or insensitivity of each metric to different underlying properties.
To create corrupted versions of the GT, we select either a subset of the GT table's rows or a subset of its columns, where each row or each column from the GT is selected with probability $\textrm{p}(x)$, preserving their original order and discarding the rest.

For the experiments we use tables from the test set of the PubTables-1M dataset, which provides text content and location information for every cell, including blank cells.
To make sure each remaining grid cell has well-defined content and location after removing a subset of rows and columns from a table, we use the 44,381 tables that do not have any spanning cells.
In order to make the metrics more comparable, we define a version of TEDS that removes functional analysis from the evaluation, called $\text{TEDS}_\text{Con}$, by removing all header information in the ground truth.

In the first experiment, the goal is to test for each metric if rows and columns are given equal importance and if every cell is credited equally regardless of its absolute position.
To test these, we create missing columns or missing rows in a predicted table according to three different selection schemes.
In the first part of the experiment we select each row with probability 0.5 using the following three different selection schemes:
\begin{itemize}
    \item First: select the first 50\% of rows.
    \item Alternating: select either every odd-numbered row or every even-numbered row.
    \item Random: select 50\% of rows within a table at random.
\end{itemize}
In the second part of the experiment, we select columns using the same three schemes as for rows.
In all six cases, exactly half of the cells are missing in a predicted table whenever the table has an even number of rows and columns.

We compare the impact of each selection scheme on the metrics in \cref{fig:metrics2}.
For a metric to give equal credit to rows and columns, it should be insensitive to (produce the same value) whether half the rows are missing or half the columns are missing.
For a metric to give equal credit to each cell regardless of absolute position, it should be insensitive to which row or column the missing cell occurs in.
The results show that $\text{DAR}_\text{Con}$ is sensitive both to whether rows or columns are selected and to which rows or columns are selected.
$\text{TEDS}_\text{Con}$ is sensitive to whether rows are selected or columns are selected, but is not sensitive to which rows or columns are selected.
On the other hand, $\text{GriTS}_\text{Con}$ produces a nearly identical value no matter which scheme is used to select half of the rows or half of the columns.

In the second experiment, we select rows and columns randomly, but vary the probability $\text{p}(x)$ from 0 to 1.
We also expand the results to include all three GriTS metrics.
We show the results of this experiment in \cref{fig:metrics}.
Like in the first experiment, $\text{DAR}_\text{Con}$ produces a similar value when randomly selecting rows or randomly selecting columns, and we see that this holds for all values of $\text{p}(x)$.
Likewise, this is true not just for $\text{GriTS}_\text{Con}$ but all GriTS metrics.
On the other hand, for $\text{TEDS}_\text{Con}$, we see that the metric has a different sensitivity to randomly missing columns than to randomly missing rows, and that the relative magnitude of this sensitivity varies as we vary $\text{p}(x)$.

In \cref{fig:metrics3}, we further split the results for DAR and GriTS by their precision and recall values.
Here we see that for GriTS, not only are rows and columns equivalent, but recall and precision match the probability of rows and columns being in the prediction and ground truth, respectively.
On the other hand, DAR has a less clear interpretation in terms of precision and recall.
Further, for DAR we notice that there is a slight sensitivity that shows up to the choice of rows versus columns for precision, which was not noticeable when considering F-score.
Overall these results show that GriTS closely resembles the ideal metric for TSR and exhibits more desirable behavior than prior metrics for this task.

\section{Conclusion}

In this paper we introduced \textit{grid table similarity} (GriTS), a new class of evaluation metric for table structure recognition (TSR).
GriTS unifies all three perspectives of the TSR task within a single class of metric and evaluates model predictions in a table's natural matrix form.
As the foundation for GriTS, we derived a similarity measure between matrices by generalizing the two-dimensional largest common substructure problem, which is NP-hard, to 2D most-similar substructures (2D-MSS).
We proposed a polynomial-time heuristic that produces both an upper and a lower bound on the true similarity and we showed that in practice these bounds are tight, nearly always guaranteeing the optimal solution.
We compared GriTS to other metrics and demonstrated using a large dataset that GriTS exhibits more desirable behavior for table structure recognition evaluation.
Overall, we believe these contributions improve the interpretability and stability of evaluation for table structure recognition and make it easier to compare results across different types of modeling approaches.

\section{Acknowledgements}
We would like to thank Pramod Sharma, Natalia Larios Delgado, Joseph N. Wilson, Mandar Dixit, John Corring, Ching Pui WAN, and the anonymous reviewers for helpful discussions and feedback while preparing this manuscript.

\bibliographystyle{splncs04}  
\bibliography{references}

\end{document}